\documentclass[journal,twoside,web]{ieeecolor}
\usepackage[table]{xcolor}
\usepackage{amsmath,amssymb,amsfonts}
\usepackage{graphicx}
\usepackage{algorithm,algorithmic}
\usepackage{multirow}
\usepackage{textcomp}
\usepackage{pifont}
\usepackage{cite}
\usepackage{soul}
\usepackage{tabularx}
\usepackage{xcolor}
\usepackage{tikz}
\definecolor{orcidgreen}{HTML}{A6CE39}

\newcommand{\orcidicon}[1]{%
  \textsuperscript{%
    \href{https://orcid.org/#1}{%
      \begin{tikzpicture}[baseline=-0.1em]
        \fill[orcidgreen] (0,0) circle (1.5ex);
        \node[white, scale=0.8, font=\bfseries\sffamily] at (0,0) {iD};
      \end{tikzpicture}%
    }%
  }%
}

\makeatletter
\let\NAT@parse\undefined
\makeatother

\usepackage[colorlinks=true, linkcolor=blue, citecolor=blue, urlcolor=blue]{hyperref}

\begin{document}

\title{Preparation of Papers for IEEE {\sc Transactions on Image Processing} (April 2025)} 

\author{Debopom Sutradhar \orcidicon{0009-0008-3811-0228}, Md. Abdur Rahman \orcidicon{0009-0004-3097-8576}, Mohaimenul Azam Khan Raiaan \orcidicon{0009-0006-4793-5382}, Reem E. Mohamed \orcidicon{0000-0002-6992-6608}, Sami Azam \orcidicon{0000-0001-7572-9750}
\thanks{Manuscript Submitted 19\textsuperscript{th} April, 2025}
\thanks{(Corresponding author: Sami Azam) }
\thanks{This work did not involve human subjects or animals in its research.}
\thanks{Debopom Sutradhar, and Md. Abdur Rahman contributed equally, and they are affiliated with the Department of Computer Science and Engineering, United International University, Dhaka, Bangladesh,  along with Mohaimenul Azam Khan Raiaan (e-mail: dsutradhar201046@bscse.uiu.ac.bd, mrahman202260@bscse.uiu.ac.bd, mraiaan191228@bscse.uiu.ac.bd).}
\thanks{Reem E. Mohamed is with the Faculty of Science and Information Technology, Charles Darwin University, Sydney, NSW, Australia (e-mail:  reem.sherif@cdu.edu.au).}
\thanks{Sami Azam is with the Faculty of Science and Technology, Charles Darwin University, Casuarina, NT 0909, Australia (e-mail:  sami.azam@cdu.edu.au).}}


\title{A Source-Free Approach for Domain Adaptation via Multiview Image Transformation and Latent Space Consistency}

\markboth{IEEE Transactions on Image Processing, Vol. 00, No. 0, April 2025}
{Sutradhar \MakeLowercase{\textit{et al.}}: IEEE Transactions on Image Processing}

\maketitle

\begin{abstract}
Domain adaptation (DA) addresses the challenge of transferring knowledge from a source domain to a target domain where image data distributions may differ. Existing DA methods often require access to source domain data, adversarial training, or complex pseudo-labeling techniques, which are computationally expensive. To address these challenges, this paper introduces a novel source-free domain adaptation method. It is the first approach to use multiview augmentation and latent space consistency techniques to learn domain-invariant features directly from the target domain. Our method eliminates the need for source-target alignment or pseudo-label refinement by learning transferable representations solely from the target domain by enforcing consistency between multiple augmented views in the latent space. Additionally, the method ensures consistency in the learned features by generating multiple augmented views of target domain data and minimizing the distance between their feature representations in the latent space. We also introduce a ConvNeXt-based encoder and design a loss function that combines classification and consistency objectives to drive effective adaptation directly from the target domain. The proposed model achieves an average classification accuracy of 90. 72\%, 84\%, and 97. 12\% in Office-31, Office-Home and Office-Caltech datasets, respectively. Further evaluations confirm that our study improves existing methods by an average classification accuracy increment of +1.23\%, +7.26\%, and +1.77\% on the respective datasets.
\end{abstract}



\section{Introduction}
\IEEEPARstart{I}{mage} classification models often assume that training and testing data share the same underlying distribution. However, in real-world applications, variations in imaging conditions frequently result in domain shifts. These shifts degrade the performance of models trained in a single source domain when deployed on different target domains. Domain adaptation (DA) addresses this challenge by enabling models to generalize across domains with differing distributions.

Recent advances in DA focus on aligning image feature distributions to mitigate domain shifts between source and target domains. A particularly promising method, source-free domain adaptation (SFDA), eliminates the need for source domain data during the training process. SFDA techniques typically rely on curriculum learning, pseudo-labeling, and entropy minimization \cite{karim2023c} \cite{ding2022source} \cite{zhang2023class}. However, pseudo-labeling can lead to error accumulation and confirmation bias, especially when early predictions are incorrect. Similarly, curriculum learning, while effective for reducing label noise, may neglect more challenging examples, limiting the model's ability to adapt fully. Additionally, the absence of source supervision makes SFDA methods prone to catastrophic forgetting, where the model loses knowledge learned from the source domain, weakening its ability to retain transferable features. To address these challenges, curriculum learning progressively incorporates high-confidence pseudo-labels to reduce label noise, ensuring that the model adapts more effectively. Similarly, dynamic pseudo-labeling strategies and entropy-based methods are used to minimize uncertainty in generated labels to further improve adaptation performance \cite{karim2023c} \cite{ma2024source}. Additionally, memory-based methods also store class prototypes learned during source training, align target samples with these prototypes, and enable adaptation without direct reliance on source-target comparisons. 

Additionally, multi-modal and fusion strategies have been developed to handle complex environments involving diverse data modalities. Generative models, such as GANs, are used for pretraining by synthesizing realistic representations of complex data distributions. For example, GAN-based frameworks generate synthetic samples that preserve intermodal correlations, which are later fine-tuned for downstream fusion tasks \cite{piechocki2023multimodal}. These tasks often involve learning unified feature representations that can seamlessly integrate information from disparate sources. In contrast, energy-based models (EBMs) capture latent dependencies between multimodal data sources in some works \cite{cui2023learning}. These approaches aim to handle multisource and multimodal data and improve feature extraction by learning shared representations across various data types. Advanced methods often integrate attention mechanisms into these frameworks to enable the model to dynamically weigh contributions from each modality based on their relevance to the task.

 DA methods have also been extended to address scenarios with highly imbalanced data distributions. Domain-specific loss functions penalize the misclassification of underrepresented classes more heavily to ensure that the model focuses more on these difficult cases \cite{yan2024transcending}. Revisiting mechanisms adjust sample weights according to class distribution to further mitigate the impact of imbalanced data. Furthermore, regularization techniques such as entropy minimization or adversarial training are used to prevent overfitting and ensure that the model generalizes better to diverse and skewed datasets \cite{zhang2024revisiting}.

To overcome existing challenges, this study introduces a novel Source-free Domain Adaptation approach. Unlike single-view transformation strategies, our framework employs a multiview augmentation strategy to enable the model to observe diverse views of the same sample and learn consistent representations across them to identify domain-invariant features. To the best of our knowledge, we are the first to integrate multiview augmentation, latent space consistency, and source-free operation into a unified framework that learns domain-invariant features directly from the target domain without relying on source supervision. Our method further eliminates reliance on adversarial setups and pseudo-labeling as well. The major contributions of our study are as follows.
\begin{itemize}
    \item We propose a novel multiview augmentation strategy that generates multiple augmented views for each image in the target domain. Random transformations are applied to simulate different imaging conditions to help the model learn domain-invariant features without requiring source data.
    \item We introduce a novel latent space consistency mechanism in which latent feature representations of augmented views are aligned in the latent space. The model learns consistent features and addresses domain shifts by minimizing the distance between augmented view representations.
    \item An objective function is incorporated that combines classification and consistency losses to retain discriminative power for target domain samples while ensuring feature consistency across augmented views.
    \item A ConvNeXt-based encoder is used for its feature extraction capabilities via depthwise convolutions to map the input data to a latent space for domain adaptation without adversarial training. The encoder is customized to reduce dimensionality and focus on essential features for domain adaptation.
\end{itemize}

The remainder of this paper is organized as follows. Section \ref{lit_rev} reviews recent related work in domain adaptation. Section \ref{method} presents our proposed methodology, including problem formulation and model architecture. Section \ref{exp_setup} provides experimental setup and training details. Section \ref{results} details the analysis of the results, performance comparisons, and ablation studies. Section \ref{discussion} provides a discussion of the study, including limitations and potential future improvements. Finally, Section \ref{conc} concludes the paper by summarizing our key findings.

\section{Related Works}
\label{lit_rev}
This section provides a review of recent domain adaptation methods, which we have grouped into four primary categories: pseudolabeling-based techniques, representation learning methods, adversarial and contrastive learning approaches, and multiview augmentation techniques.

\vspace{0.02\linewidth}\noindent\textbf{\textit{Pseudo-labeling}} is an effective technique in the adaptation of the source-free domain, where the provisional labels assigned to the target samples guide the refinement of the features and the alignment of the domain using the predictions of the model to iteratively adapt to the target distribution without requiring access to source data. Recent advances \cite{liang2020we,yang2021exploiting,tang2022semantic,litrico2023guiding,tang2024source} have introduced innovative refinements to enhance the effectiveness of pseudo-labeling. For example, Liang et al. \cite{liang2020we} introduced SHOT, which focuses on aligning target features through information maximization and self-supervised learning. Yang et al. \cite{yang2021exploiting} introduced NRC, which refines target representations through consistency based on clustering and neighborhood topology. Similarly, Tang et al. \cite{tang2022semantic} developed SCLIM, using enhanced k-means clustering and semantic neighbor topology to improve adaptation. In contrast, Litrico et al. \cite{litrico2023guiding} presented PLUE, which incorporates uncertainty-based loss reweighting to enhance the reliability of pseudo-labels. Additionally, Tang et al. \cite{tang2024source} proposed TPDS, a method that aligns target features through proxy distributions and manifold geometry. These methods show a steady improvement in the average classification accuracy (in multiple domains) on the Office-Home dataset, with SHOT achieving 72.3\% and TPDS reaching 79.5\%. When comparing these approaches, methods that combine clustering and uncertainty estimation, such as PLUE (77.6\%) and TPDS (79.5\%), tend to outperform simpler techniques such as SHOT and NRC, particularly when adapting to more complex domain shifts.

\vspace{0.02\linewidth}\noindent\textbf{\textit{Representation Learning}} approaches focus on extracting robust and transferable features that reduce domain discrepancies. Recent works \cite{ding2022source, tang2024source1, jin2020minimum, zhu2024multiview} have advanced this area by introducing innovative strategies for feature alignment and domain adaptation. For example, Tang et al. \cite{tang2024source1} introduced DIFO, which combines pre-trained vision-language models (e.g., CLIP) with task-specific prompt learning to effectively fine-tune and align features. In another study, Ding et al. \cite{ding2022source} introduced SFDA-DE, which employs spherical k-means clustering to simulate source-like features and estimate class-conditioned distributions. In a different approach, Jin et al. \cite{jin2020minimum} developed MCC, a method that introduces a loss function to reduce class confusion and improve generalizability of learned representations. Zhu et al. \cite{zhu2024multiview} presented a multiview latent space framework (MSL) that refines pseudo-label selection by integrating raw, fine-tuned, and pseudo-labeled features. These methods report competitive average accuracy results across datasets, with MCC, SFDA-DE, and DIFO achieving accuracies between 83\% and 90.1\% in Office-31, while MSL reaches 96.8\% in Office-Caltech.

\begin{figure*}
\centering
\includegraphics[scale=0.17]{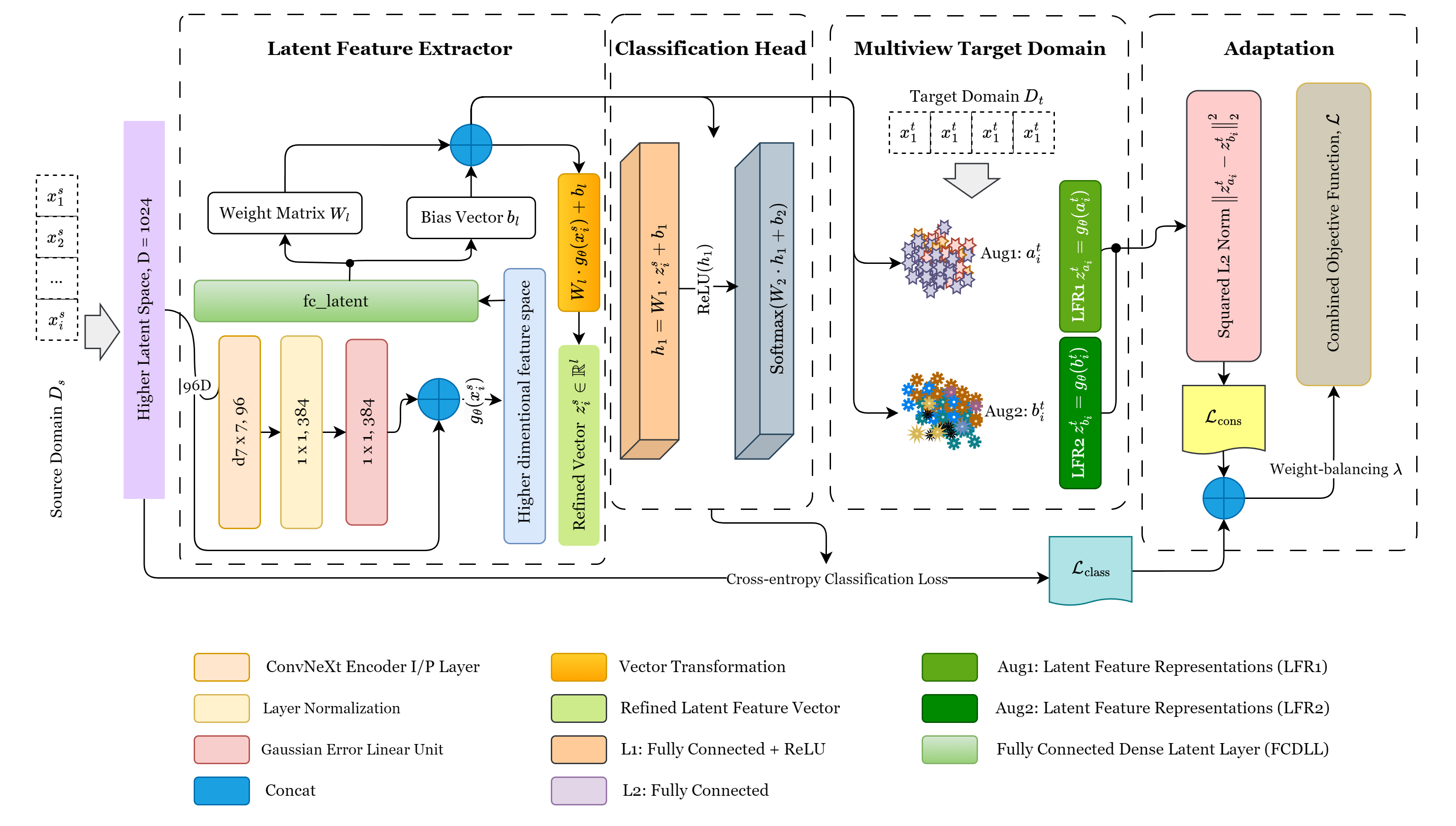} 
\caption{Proposed methodology illustrating the latent feature extraction, classification, multiview representation with augmentations and adaptation with consistency loss for domain adaptation.}
\label{fig:fig_1}
\end{figure*}

\vspace{0.02\linewidth}\noindent\textbf{\textit{Adversarial and Contrastive Learning Frameworks}} aim to enforce domain invariance by aligning source and target distributions or preserving target structure. Approaches such as \cite{cui2020gradually, kang2019contrastive,tang2020unsupervised,xia2021adaptive} exemplify this area. For example, Cui et al. \cite{cui2020gradually} proposed GVB-GD, which effectively uses a gradually vanishing bridge to stabilize adversarial training by gradually decreasing the influence of the generator. Similarly, Kang et al. \cite{kang2019contrastive}  introduced CAN, which aligns intra-class and inter-class discrepancies through contrastive learning. On the other hand, Tang et al. \cite{tang2020unsupervised} applied structural regularization in SRDC to improve target discrimination, making the target features more distinct and aligned with the source domain. In another work, Xia et al. \cite{xia2021adaptive} combined adversarial inference with category-wise matching in A²Net, which adapts more robustly to target domains by distinguishing between source-similar and source-dissimilar samples. On Office-31, these methods achieve average accuracies ranging from 89.3\% (GVB-GD) to 90.8\% (SRDC), with structurally guided approaches such as SRDC outperforming (90.8\%) purely adversarial methods.
\vspace{0.02\linewidth}\noindent\textbf{\textit{Multiview Augmentation}} is often used for resilient domain adaptation where source data is unavailable \cite{orbes2019multi, orbes2022augmentation, zhu2024multiview, tran2024multi}. For example, in a study, Orbes-Arteaga et al. \cite{orbes2019multi} proposed a multidomain adaptation method for MRI segmentation of the brain. They used paired consistency between different types of MRI scan as a form of discrete augmentation and combined it with adversarial learning to achieve robust segmentation across multiple target domains. This approach highlights the effectiveness of using natural multiview data generated from different MRI sequences for domain adaptation. In another study, Xie et al. \cite{xie2020unsupervised} explored unsupervised data augmentation for consistency training in semi-supervised learning. Although not specifically focused on domain adaptation, their work demonstrated that sophisticated enhancement strategies can significantly improve model performance by generating diverse views of the data to enhance robustness and generalization. These findings motivated our proposed approach for source-free domain adaptation, where augmentation can be applied only to the target domain to simulate the source domain.

Despite significant advancements in domain adaptation, existing methods face several limitations. For instance, pseudo-labeling techniques often struggle with unreliable label assignments, especially under complex domain shifts, which can propagate errors and hinder effective adaptation. Representation learning approaches frequently require extensive fine-tuning or rely on pre-trained models. Adversarial and contrastive learning frameworks tend to be computationally intensive and depend on adversarial training. Moreover, many methods require access to source domain data during adaptation, which restricts their applicability. In this study, we address these issues by introducing a novel domain adaptation method.

\section{Methodology}
\label{method}
\subsection{Problem Objective}
The objective is to learn a model \(f_{\theta}: \mathcal{X} \to \mathcal{Y}\) that performs well on the target data \(D_{t}\) without requiring access to the source data \(D_{s}\) during adaptation. To achieve this, we create multiple augmented views \(a_{i}^{t},\ b_{i}^{t}\) for each \(x_{i}^{t} \in D_{t}\). For each pair \(\left( a_{i}^{t}, b_{i}^{t} \right)\), we enforce a latent consistency constraint by minimizing a distance metric \(d\left( f_{\theta}\left( a_{i}^{t} \right), f_{\theta}\left( b_{i}^{t} \right) \right)\) in the latent space. This constraint encourages \(f_{\theta}\) to learn domain-invariant features by aligning representations across augmented views. Finally, the overall objective function combines a classification loss \(L_{\text{class}}\) with a consistency loss \(L_{\text{cons}}\). Fig. \ref{fig:fig_1} illustrates the overall pipeline of the proposed methodology.

\subsection{Initial Training}
In this phase, the goal is to train a feature extractor and classifier that can effectively map input samples to their corresponding class labels. We begin with each dataset \(D = \left( x_{i}^{s},\ y_{i}^{s} \right)\), where \(\ x_{i}^{s} \in Y\ \)represents input samples, and \(y_{i}^{s} \in \ Y\)are the corresponding one-hot encoded labels. Our architecture uses ConvNeXt as the backbone of an encoder \(g_{\theta}\) to extract hierarchical feature representations from \(x_{i}^{s}\). ConvNeXt is a convolutional neural network designed to enhance feature extraction through depthwise convolutions, Layer Normalization (LN), and GELU activations \cite{liu2022convnet}. The encoder is composed of stacked convolutional blocks, each of which processes spatial and semantic information in the input. For a given input \(x_{i}^{s}\), the encoder produces a high-dimensional representation \(g_{\theta}(x_{i}^{s}) \in \mathbb{R}^{d}\). To refine these features, we incorporate a fully connected layer (fc\_latent), which projects the high-dimensional representation, \(g_{\theta}(x_{i}^{s})\), into a latent space of reduced dimensionality \(l\), defined in Equation \eqref{eq:eq_1} as:
\begin{equation}
    z_{i}^{s}\  = \ W_{l}\  \bullet \ g_{\theta}\left( x_{i}^{s} \right) + \ b_{l}
\label{eq:eq_1}
\end{equation}

where \(W_{l} \in \mathbb{R}^{l \times d}\) and \(b_{l} \in \mathbb{R}^{l}\) are the weight matrix and bias vector, respectively. The latent vector \(z_{i}^{s} \in \mathbb{R}^{l}\) encapsulates discriminative features optimized for classification. The latent embedding \(z_{i}^{s}\) then passes through the classification head \(h_{\theta}\), which consists of a fully connected two-layer network with ReLU activation. The classification logits are computed as shown in Equation \eqref{eq:eq_2}, and \eqref{eq:eq_3}:
\begin{equation}
    z^{hidden}\  = \ \sigma\left( W_{1} \bullet z_{i}^{s}\  + \ b_{1} \right)
\label{eq:eq_2}
\end{equation}

\begin{equation}
    {\widehat{y}}_{i}^{s}\  = \ softmax\left( W_{2} \bullet z^{hidden}\  + \ b_{2} \right)\
\label{eq:eq_3}
\end{equation}

where \(W_{1} \in \mathbb{R}^{h \times l}\), \(W_{2} \in \mathbb{R}^{c \times l}\), and \(b_{1} \in \mathbb{R}^{h}\), \(b_{2} \in \mathbb{R}^{c}\) are the learnable parameters; \(C\) and \(\sigma\) are the number of classes and activation function, respectively. Following this, we employ a cross-entropy classification loss for accurate class separation, defined in Equation 4:
\begin{equation}
\ \ \mathcal{L}_{\text{class\ }} = - \frac{1}{N_{s}}\sum_{i = 1}^{N_{A}}\mspace{2mu}\sum_{c = 1}^{C}\mspace{2mu} y_{i,c}^{s}\log h_{\theta}\left( z_{i}^{s} \right)\
\label{eq:eq_4}
\end{equation}

where \(C\) is the number of classes, \(\ y_{\left\{ i,c \right\}}^{s}\) is the one-hot encoded label for class \(c\), and \(h_{\theta}\left( z_{i}^{s} \right)\) is the predicted probability for class \(c\). This objective guides the encoder \(g_{\theta}\) to form distinct class boundaries in the latent space.

\subsection{Latent Space Consistency with Multiview Augmentation}
To maintain transferable feature representations, we impose latent space consistency across augmented views of the target samples. For each target sample \(x_{i}^{t}\), we generate two augmented views, \(\ a_{i}^{t}\) and \(b_{i}^{t}\text{.}\) Through transformations such as random rotations, color jitter, and resizing, these augmentations simulate different environmental imaging conditions. We pass both augmented views through the same encoder \(g_{\theta}\). The encoder produces two latent feature representations, \(\ z_{a_{i}^{t}} = g_{\theta}\ \left( a_{i}^{t} \right)\) and \(z_{b_{i}^{t}} = g_{\theta}\ \left( b_{i}^{t} \right)\), for each augmented view. The goal is to make these representations as similar as possible in the latent space. To enforce this, we minimize the loss of latent space consistency between the augmented view representations. The equation is represented by Equation \eqref{eq:eq_5}:
\begin{equation}
    \mathcal{L}_{cons} = \frac{1}{N_{t}}\sum_{i = 1}^{N_{t}}\mspace{2mu}\left\| z_{a_{i}}^{t} - z_{b_{i}}^{t} \right\|_{2}^{2}\
\label{eq:eq_5}
\end{equation}

where \(N_{t}\) is the number of target domain samples, and \(\| \cdot \|_{2}^{2}\) refers to the squared L2 distance, which measures how far apart two vectors are. \(z_{a_{i}}^{t}\) and \(z_{b_{i}}^{t}\) are the latent features (embeddings) generated by the encoder \(g_{\theta}\) from two different augmented versions of the same target sample, \(a_{i}^{t}\) and \(b_{i}^{t}\), respectively. The L2 norm is used to penalize large differences in feature vectors directly, which promotes a tight alignment of augmented view representations. The loss \(\mathcal{L}_{cons}\) encourages the encoder to generate similar latent embeddings for the augmented views of the same target sample, ensuring consistent representations across different augmentations.

\subsection{Combined Objective Function for Classification and Consistency}
The overall objective function integrates classification and consistency constraints. The combined objective helps \(f_{\theta}\) retain transferable features while adapting to domain changes in target domains \(D_{t}\). Balances source-specific class features with target consistency. The combined loss is calculated as shown in Equation \eqref{eq:eq_6}:
\begin{equation}
    \mathcal{L =}\mathcal{L}_{\text{class\ }} + \lambda\mathcal{L}_{\text{cons\ }}\
\label{eq:eq_6}
\end{equation}

where \(\lambda\) is introduced to balance the two objectives of the loss function: maintaining discriminative features learned from input data and enforcing consistency in augmented views in the target domain. The optimal value of \(\lambda\) was determined by empirical tuning (see Section \ref{ablation}). Here, \(\mathcal{L}_{\text{class\ }}\) retains the class boundaries from the source phase, while \(\mathcal{L}_{\text{cons\ }}\) adapts to the target variations. Both are summed to allow simultaneous optimization of classification accuracy and feature alignment, which ensures that both objectives contribute proportionally to the overall loss.

This combined objective is used to guide the encoder \(g_{\theta}\) and the classifier \(h_{\theta}\) to jointly minimize the classification error in the source domain and maximize the consistency in the target domain. Unlike traditional adversarial or pseudo-labeling methods, this approach simplifies adaptation by leveraging inherent invariance in the feature space.


\section{Experimental Setup}
\label{exp_setup}
\subsection{Datasets}
We conducted experiments on three widely used domain adaptation benchmarks to evaluate our proposed methodology. Although DomainNet \cite{peng2019moment}, VisDA-2017 \cite{peng2017visda}, or PACS \cite{li2017deeper} are used in domain adaptation as benchmarks, they are computationally demanding. This makes them less ideal for real-world applications where efficiency is a concern. Thus, we conducted experiments on three widely used domain adaptation benchmarks to evaluate our proposed methodology, which are Office-31, Office-home, and Office-Caltech. These datasets are not as computationally demanding as DomainNet, VisDA-2017, or PACS. They also represent varying levels of domain shift difficulty, where Office-31 has small shifts, while Office-Home has larger shifts. Furthermore, they allow for a direct comparison with previous SFDA methods, ensuring fair benchmarking.

\vspace{0.02\linewidth}\noindent\textbf{Office-31 \cite{saenko2010adapting}:} Office-31 is a small-scale benchmark with three domains: Amazon (2,817 images), DSLR (498 images), and Webcam (795 images). It contains a total of 4,110 images belonging to 31 categories, collected from real-world scenarios. In addition, it is one of the most commonly used benchmarks, enabling direct comparisons with previous works. 

\vspace{0.02\linewidth}\noindent\textbf{Office-Home \cite{venkateswara2017deep}:} Office-Home is a large-scale benchmark dataset comprising four visually dissimilar domains: Artistic images, Clipart images, Product images, and Real-world images. This dataset includes 12 transfer tasks and contains a total of 15,500 images from 65 categories.  
\begin{figure}
\centering
\includegraphics[scale=0.4]{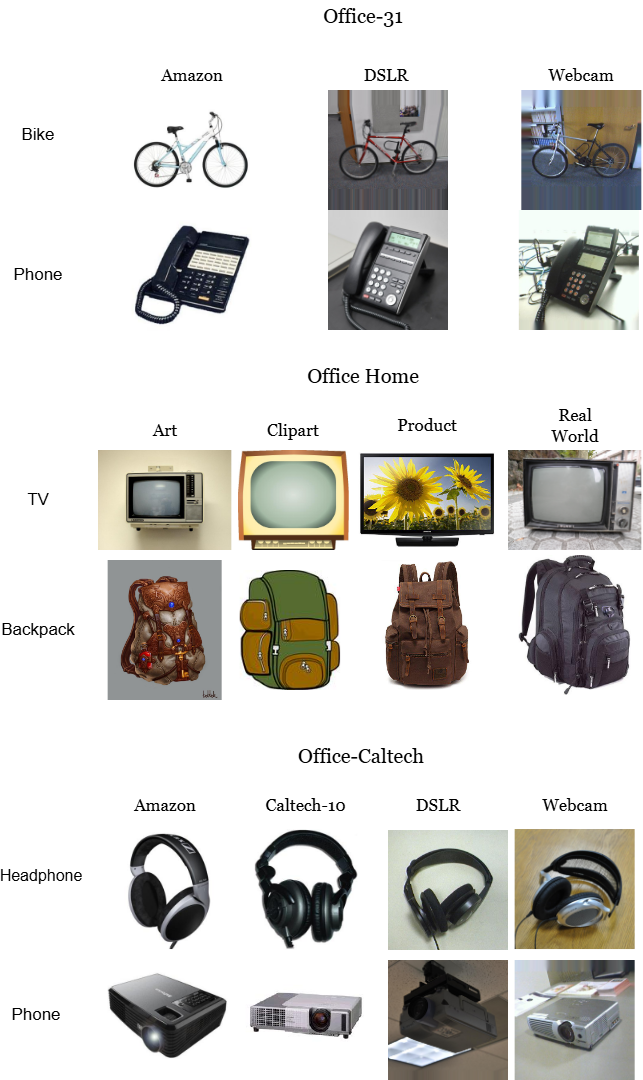} 
\caption{Sample images from the datasets.}
\label{fig:fig_2}
\end{figure}

\vspace{0.02\linewidth}\noindent\textbf{Office-Caltech \cite{gong2012geodesic}:} Office-Caltech is a standardized benchmark for domain adaptation. It combines the Office-10 and Caltech-10 datasets, which share 10 common object categories. They range greatly in visual attributes such as resolution, background clutter, and lighting conditions. This controlled overlap minimizes class-level confounding effects and enables a thorough evaluation of a model's generalization across different domains.
The sample photos from various datasets, as shown in Fig. \ref{fig:fig_2}, illustrate the variety of the different classes in different domains.

\begin{table*}[]
\centering
\large
\renewcommand{\arraystretch}{1.7}
\caption{Classification accuracies (\%) on the Office-Home dataset across Art (Ar), Clipart (Cl), Product (Pr), and Real-World (Rw) domains. The highest and the second highest performances are highlighted in bold and underscore, respectively.}
\label{table_1}
\resizebox{\textwidth}{!}{%
\begin{tabular}{lcccccccccccccccc}
\hline
\textbf{Method} & \textbf{SF} & \textbf{Ar→Cl} & \textbf{Ar→Pr} & \textbf{Ar→Rw} & \textbf{Cl→Ar} & \textbf{Cl→Pr} & \textbf{Cl→Rw} & \textbf{Pr→Ar} & \textbf{Pr→Cl} & \textbf{Pr→Rw} & \textbf{Rw→Ar} & \textbf{Rw→Cl} & \textbf{Rw→Pr} & \textbf{Min.} & \textbf{Max.} & \textbf{Avg.} \\
\hline
SHOT \cite{liang2020we}      & \textcolor{green}{\checkmark} & 56.7 & 77.7 & 84.6 & 63.2 & 78.8 & 79.4 & 67.9 & 54.5 & 82.3 & 74.2 & 66.8 & 82.8 & 54.5 & 82.8 & 72.3 \\
NRC \cite{yang2021exploiting}      & \textcolor{green}{\checkmark} & 56.6 & 82.2 & 84.3 & 66.6 & 82.4 & 83.6 & 69.3 & 58.4 & 83.6 & 76.8 & 65.3 & 85.2 & 56.6 & 85.2 & 74.0 \\
SCLIM \cite{tang2022semantic}    & \textcolor{green}{\checkmark} & 58.4 & 79.7 & 83.6 & 67.3 & 81.2 & 82.5 & 70.8 & 59.5 & 84.2 & 76.8 & 64.8 & 86.3 & 58.4 & 86.3 & 74.2 \\
ELR \cite{yi2023source}      & \textcolor{green}{\checkmark} & 57.1 & 79.9 & 83.4 & 67.6 & 82.0 & 82.9 & 70.6 & 58.6 & 83.4 & 76.3 & 65.2 & 85.4 & 57.1 & 85.4 & 73.7 \\
PLUE \cite{litrico2023guiding}     & \textcolor{green}{\checkmark} & 63.4 & 83.3 & 85.8 & 73.8 & 83.4 & 85.6 & 74.2 & 63.4 & 86.0 & 80.4 & 68.3 & 87.2 & 63.4 & 87.2 & 77.6 \\
TPDS \cite{tang2024source}     & \textcolor{green}{\checkmark} & 59.3 & 80.3 & 82.1 & 70.6 & 79.4 & \underline{80.9} & 69.8 & 56.8 & \underline{82.1} & \underline{74.5} & 61.2 & 85.3 & 59.3 & 85.3 & 73.5 \\  
DIFO-C-RN \cite{tang2024source1} & \textcolor{green}{\checkmark} & 62.6 & 87.3 & 87.1 & 79.5 & 86.9 & 88.1 & 78.6 & 66.2 & 88.9 & 82.8 & 63.3 & 87.7 & 62.6 & 88.9 & 79.5 \\
DIFO-C-B32 \cite{tang2024source1} & \textcolor{green}{\checkmark} & \textbf{70.6} & \underline{90.6} & \underline{88.8} & \underline{82.5} & \textbf{90.6} & \textbf{88.9} & \underline{80.9} & \underline{70.1} & \underline{88.9} & \textbf{83.4} & \textbf{70.5} & \underline{91.2} & \underline{70.1} & \underline{91.2} & \underline{83.1} \\
\cellcolor{gray!20}OURS          & \cellcolor{gray!20}\textcolor{green}{\checkmark} & \cellcolor{gray!20}\underline{67.7} & \cellcolor{gray!20}\textbf{92.4} & \cellcolor{gray!20}\textbf{90.1} & \cellcolor{gray!20}\textbf{85.2} & \cellcolor{gray!20}\underline{89.7} & \cellcolor{gray!20}\textbf{89.9} & \cellcolor{gray!20}\textbf{81.2} & \cellcolor{gray!20}\textbf{75.2} & \cellcolor{gray!20}\textbf{90.7} & \cellcolor{gray!20}81.4 & \cellcolor{gray!20}\underline{69.5} & \cellcolor{gray!20}\textbf{94.5} & \cellcolor{gray!20}\textbf{67.7} & \cellcolor{gray!20}\textbf{94.5} & \cellcolor{gray!20}\textbf{84.0} \\
\hline
\end{tabular}%
}
\end{table*}

\begin{table*}[]
\centering
\renewcommand{\arraystretch}{1.3}
\caption{Classification accuracies (\%) on the Office-31 dataset across Amazon (A), DSLR (D), and Webcam (W) domains.}
\label{table_2}
\scriptsize
\begin{tabular}{lcccccccccc}
\hline
\textbf{Method} & \textbf{SF} & \textbf{A→D} & \textbf{A→W} & \textbf{D→A} & \textbf{D→W} & \textbf{W→A} & \textbf{W→D} & \textbf{Min.} & \textbf{Max.} & \textbf{Avg.} \\
\hline
MDD \cite{zhang2019bridging}      & \ding{55} & 93.5 & 94.5 & 74.6 & 98.4 & 72.2 & \textbf{100}  & 72.2 & \textbf{100} & 88.9 \\
GVB-GD \cite{cui2020gradually}   & \ding{55} & 95.0 & 94.8 & 73.4 & 98.7 & 73.7 & \textbf{100} & 73.4 & \textbf{100} & 89.3 \\
MCC \cite{jin2020minimum}      & \ding{55} & \underline{96.5} & \underline{95.4} & 72.6 & 98.6 & 73.9 & \underline{99.9} & 72.6 & \underline{99.9} & 89.5 \\
GSDA \cite{hu2020unsupervised}     & \ding{55} & 94.8 & \textbf{95.7} & 73.7 & \underline{99.1} & 74.9 & 99.8 & 73.7 & 99.8 & 89.7 \\
CAN \cite{kang2019contrastive}      & \ding{55} & 95.0 & 94.5 & 78.0 & \underline{99.1} & 77.0 & 99.8 & \underline{77.0} & 99.8 & 90.6 \\
SRDC \cite{tang2020unsupervised}     & \ding{55} & 95.8 & \textbf{95.7} & \underline{76.7} & \textbf{99.2} & 77.1 & \textbf{100}  & \textbf{77.1} & \textbf{100} &  \textbf{90.8} \\
SFDA \cite{kim2021domain}     & \textcolor{green}{\checkmark} & 92.2 & 91.1 & 71.0 & 98.2 & 71.2 & 99.5 & 71.0 & 99.5 &  87.2 \\
SHOT \cite{liang2020we}      & \textcolor{green}{\checkmark} & 93.8 & 94.1 & 74.7 & 98.4 & 74.3 & \underline{99.9} & 74.3 & \underline{99.9} &  88.6 \\
3C-GAN \cite{li2020model}   & \textcolor{green}{\checkmark} & 92.7 & 93.2 & 75.3 & 98.5 & \textbf{77.8} & 99.8 & 75.3 & 99.8 &  89.6 \\
A²Net \cite{xia2021adaptive}    & \textcolor{green}{\checkmark} & 94.5 & 94.0 & \underline{76.7} & 99.2 & \underline{76.1} & \textbf{100} & 76.1 & \textbf{100 }& 90.1 \\
SFDA-DE \cite{ding2022source}   & \textcolor{green}{\textcolor{green}{\checkmark}} & 96.0 & 94.2 & 76.6 & 98.5 & 75.5 & 99.8 & 75.5 & 99.8 & 90.1 \\
\cellcolor{gray!20}OURS          & \cellcolor{gray!20}\textcolor{green}{\checkmark} & \cellcolor{gray!20}\textbf{97.99} & \cellcolor{gray!20}95.2 & \cellcolor{gray!20}\textbf{76.9} & \cellcolor{gray!20}\textbf{99.2} & \cellcolor{gray!20}76.0 & \cellcolor{gray!20}99.0 & \cellcolor{gray!20}76.0 & \cellcolor{gray!20}99.2  & \cellcolor{gray!20}\underline{90.72} \\
\hline
\end{tabular}%
\end{table*}
\subsection{Model Training}
Our model is trained using a self-supervised multiview consistency strategy to adapt to the target domain. The backbone is a pre-trained ConvNeXt-B network, which we have modified by adding a latent projection layer and a classification head. Two augmented views are generated for each target domain sample during training using separate augmentation pipelines. The first augmentation pipeline introduces appearance-level changes through random horizontal flips (p = 0.5) and color jittering with brightness, contrast, and saturation set to 0.4, and hue variation of 0.1. The second pipeline captures geometric transformations via random rotation (±20°) and random resized cropping to 224$\times$224 with a scale range between 0.8 and 1. These augmentations are intentionally chosen to reflect natural changes in imaging conditions, such as variations in camera angle, ambient lighting, and object framing, which are commonly encountered in practical scenarios. Rather than relying on synthetic data or adversarial perturbations, our approach leverages these lightweight, interpretable transformations to expose the model to diverse views of the same instance, encouraging it to learn robust domain-invariant features directly from the target domain. These augmented views are processed through the model to produce latent representations. To adapt the model to the target domain, we then align these latent representations by minimizing a consistency loss (mean square error) between them. Simultaneously, a classification loss is applied to maintain class-discriminative features. The final objective combines these losses in a weighted manner, encouraging domain-invariant features and clear class separation. We incorporate mixed-precision training to improve computational efficiency. In particular, our approach operates in a source-free setup that relies only on data from the target domain for adaptation. This enables efficient domain adaptation without source domain data.

All experiments were carried out on the Kaggle cloud platform, using an NVIDIA Tesla T4 GPU with 16 GB of VRAM, an Intel(R) Xeon(R) CPU, and 13 GB of RAM. Model training was used with mixed precision (AMP) to optimize memory usage and computational efficiency. This setup provided sufficient performance to train the ConvNeXt-based encoder with multiview consistency across standard domain adaptation benchmarks while ensuring practical reproducibility on accessible cloud infrastructure.

\section{Results}
\label{results}

\subsection{Results Analysis}
For this study, we have used accuracy as the primary evaluation metric to enable clear interpretation and comparison with existing domain adaptation methods. 

As shown in Table \ref{table_1}, our approach performs consistently in all domain adaptation tasks in the Office-Home data set. With an average accuracy of 84\%, our approach surpasses most source-free domain adaptation techniques and shows reliable and consistent performance in all domain adaptation tasks in the Office-Home dataset.


Similarly, across the Office-31 dataset, our method achieves a higher average accuracy of 90. 72\% with an increase of + 1.23\% in accuracy compared to most existing methods, including both source-free and source-dependent approaches. Notably, it sets a new standard in these tasks by achieving the maximum accuracy in A→D (97.99\%) and performing exceptionally well in D→A (99.2\%) and D→W (99\%). Although several techniques, such as MDD, GVB-GD, and SRDC, get 100\% in W→D, they fall short in other domain shifts, leading to lower averages overall (88.9\%, 89.3\%, and 90.8\%, respectively). Our method is superior in source-free domain adaptation, as evidenced by its reliable and consistent performance on all tasks, as shown in Table \ref{table_2}.
\begin{table*}[]
\centering
\large
\renewcommand{\arraystretch}{1.5}
\caption{Classification accuracies (\%) on the Office-Caltech dataset across Caltech-10 (C), Amazon (A), DSLR (D), and Webcam (W) domains.}
\label{table_3}
\scriptsize
\begin{tabular}{lcccccccccccccc}
\hline
\textbf{Method} & \textbf{C→A} & \textbf{C→W} & \textbf{C→D} & \textbf{A→C} & \textbf{A→W} & \textbf{A→D} & \textbf{W→A} & \textbf{W→D} & \textbf{W→C} & \textbf{D→A} & \textbf{D→W} & \textbf{Min.} & \textbf{Max.} & \textbf{Avg.} \\
\hline
GSM \cite{zhang2019transductive}      & \underline{96.0} & 95.9 & 96.2 & 94.6 & 89.5 & 92.4 & 94.1 & 95.8 & 100 & 93.9 & 95.1 & 89.5 & 100 &  \textbf{98.2} \\
JGSA \cite{zhang2017joint}     & 95.1 & 97.6 & \underline{96.8} & 93.9 & 96.2 & 96.2 & \underline{95.1} & \underline{95.9} & 100 & \underline{94.0} & \underline{96.3} & \textbf{93.9} & 100 & 96.2 \\
RTN \cite{long2016unsupervised}      & 93.7 & 96.9 & 94.2 & 88.1 & 95.5 & 86.6 & 92.5 & 95.0 & 100 & 84.6 & 94.8 & 84.6 & 100 & 92.2 \\
MDDA \cite{rahman2020correlation}     & 93.6 & 95.2 & 93.4 & 89.1 & 95.7 & 86.5 & 92.6 & 94.4 & 100 & 84.7 & 95.7 & 84.7 & 100 & 93.6 \\
PICCSCS \cite{li2023pseudo}  & 93.7 & \textbf{98.3} & \textbf{97.5} & 89.2 & 98.3 & 96.8 & 89.3 & 93.9 & 100 & 93.8 & \textbf{100} & 89.3 & 100 & 95.1 \\
MSL \cite{zhu2024multiview}      & 95.9 & 97.3 & \underline{96.8} & \underline{94.7} & \textbf{99.7} & \textbf{99.4} & 93.6 & 94.8 & 100 & \underline{94.0} & 95.9 & \underline{93.6}  & 100 & 96.8 \\
\cellcolor{gray!20}OURS          & \cellcolor{gray!20}\textbf{98.6} & \cellcolor{gray!20}\textbf{98.3} & \cellcolor{gray!20}\underline{96.8} & \cellcolor{gray!20}\textbf{95.2} & \cellcolor{gray!20}\underline{98.9} & \cellcolor{gray!20}\underline{99.2} & \cellcolor{gray!20}\textbf{95.4} & \cellcolor{gray!20}\textbf{96.2} & \cellcolor{gray!20}100 & \cellcolor{gray!20}\textbf{96.5} & \cellcolor{gray!20}93.2 & \cellcolor{gray!20}93.2 & \cellcolor{gray!20}100
 & \cellcolor{gray!20}\underline{97.12} \\
\hline
\end{tabular}%

\end{table*}

Our approach also performs exceptionally well on the Office-Caltech dataset. Our method achieves remarkable performance with an average accuracy of 97.12\%, making it one of the top performing approaches overall. Although GSM outperforms us by a small margin (98. 2\%), this is due to isolated peaks in particular domain shifts. Our approach, on the other hand, continually produces notable outcomes in all tasks, highlighting its reliability and robustness. In addition, we perform best on important domain adaptation tasks like C→A, C→W, and C→D, and we do competitively on others like A→W and A→D. Even if the greatest average was not reached, this shows how well our method adapts to a variety of domains. The detailed results are shown in Table \ref{table_3}.

Table \ref{table_4} demonstrates the dominant position of our strategy in domain adaptation, with leading averages on Office-Home (84\%) and the greatest accuracy in 21 out of 30 domain shifts. It demonstrates strong, domain-invariant learning and balanced performance across shifts, as it performs exceptionally well on important Office-31 and Office-Caltech tasks.

\begin{table}[H]
\centering
\renewcommand{\arraystretch}{1.5}
\caption{Combined key results from the three datasets. The average (existing) classification accuracies (\%) on the respective datasets are 76.74\%, 89.49\%, and 95.35\%.}
\label{table_4}
\begin{tabularx}{\linewidth}{l l c c}
\hline
\textbf{Dataset}      & \textbf{Others' Best Avg.} & \textbf{Our Avg.} & \textbf{Our Highest}  \\ \hline
Office-Home           & 83.1\% (DIFO-C-B32)     & 84.00\%                & 8 out of 12           \\
Office-31             & 90.8\% (SRDC)           & 90.72\%                & 3 out of 6            \\
Office-Caltech        & 98.2\% (GSM)            & 97.12\%                & 6 out of 11           \\
\hline
\end{tabularx}
\end{table}

\subsection{Ablation Study}
\label{ablation}
We initially conducted an ablation study by experimenting with the encoder in our framework with five different backbones. VGG-19, ResNet-50, ViT, DeiT, and ConvNeXt-B. Each model was trained while keeping all hyperparameters fixed in the Office-Home data set using the Adam optimizer with a learning rate of 1$\times$10\textsuperscript{-4} (decayed by a factor of 0.1 for every 15 epochs), a batch size of 8, and a weight decay of 1$\times$10\textsuperscript{-5}, with a consistency loss weight $(\lambda)$ set to 1.0 and the same data enhancement pipeline. The results showed that VGG-19 performed the weakest at 27\% accuracy, followed by ResNet-50 at 35\%. The ViT backbone improved the accuracy to 40\%, but still had limitations that required large-scale pre-training. DeiT demonstrated a notable jump to 77\%, showing promise in domain adaptation due to its increased data efficiency. 

Ultimately, the ConvNeXt-B backbone outperformed the others with 80\% accuracy, showing its robustness for the task. Unlike ViT or DeiT, ConvNeXt retains local feature extraction through convolutions but introduces modern design principles like depthwise separable convolutions and large kernel sizes, allowing it to capture both local and global patterns effectively. With ConvNeXt-B selected as the best performing backbone, we further tuned the hyperparameters to optimize its performance. Specifically, we experimented with varying the learning rate (1$\times$10\textsuperscript{-4}, 1$\times$10\textsuperscript{-3}), adjusting the batch size (4, 8, 16, and 32), and testing different values of $\lambda$ (0.1, 0.5, 1.0, 10.0). Larger values (e.g., $(\lambda)$ = 10.0) overly constrained the latent space, leading to poor class discrimination, while smaller values (e.g. $(\lambda)$=0.1) underweighted the regularization, resulting in insufficient adaptation to domain shift \cite{sohn2020fixmatch}. After tuning, the optimal configuration achieved the highest performance with a learning rate of 1$\times$10\textsuperscript{-4}, batch size of 4, weight decay of 1$\times$10\textsuperscript{-4}, and $\lambda=0.5$. These settings consistently yielded the best trade-off between convergence speed and generalization. The tuned model achieved 83.9\% on the Office Home dataset. The settings were achieved in 90.72\% on the Office-31 dataset. This strong performance, despite the limited data setting, highlights the robust generalizability of the model. The average training time per epoch for the ConvNeXt-B backbone was approximately 2 minutes, compared to 2.9 minutes for ViT and 3.7 minutes for DeiT in identical settings (batch size of 4 and input resolution of 224$\times$224).

This two-step procedure, first identifying the most effective backbone and then fine-tuning hyperparameters, proved crucial in achieving robust performance on challenging domain shifts on the Office-Home dataset. With different experiments, we first selected our ConvNeXt-B backbone. Then we performed an extensive sensitivity analysis on key hyperparameters, including learning rate, batch size, and consistency weight ($\lambda$). By combining this hyperparameter optimization, we refined the configuration space and achieved consistent improvements. 

\begin{figure*}
\centering
\includegraphics[scale=0.47]{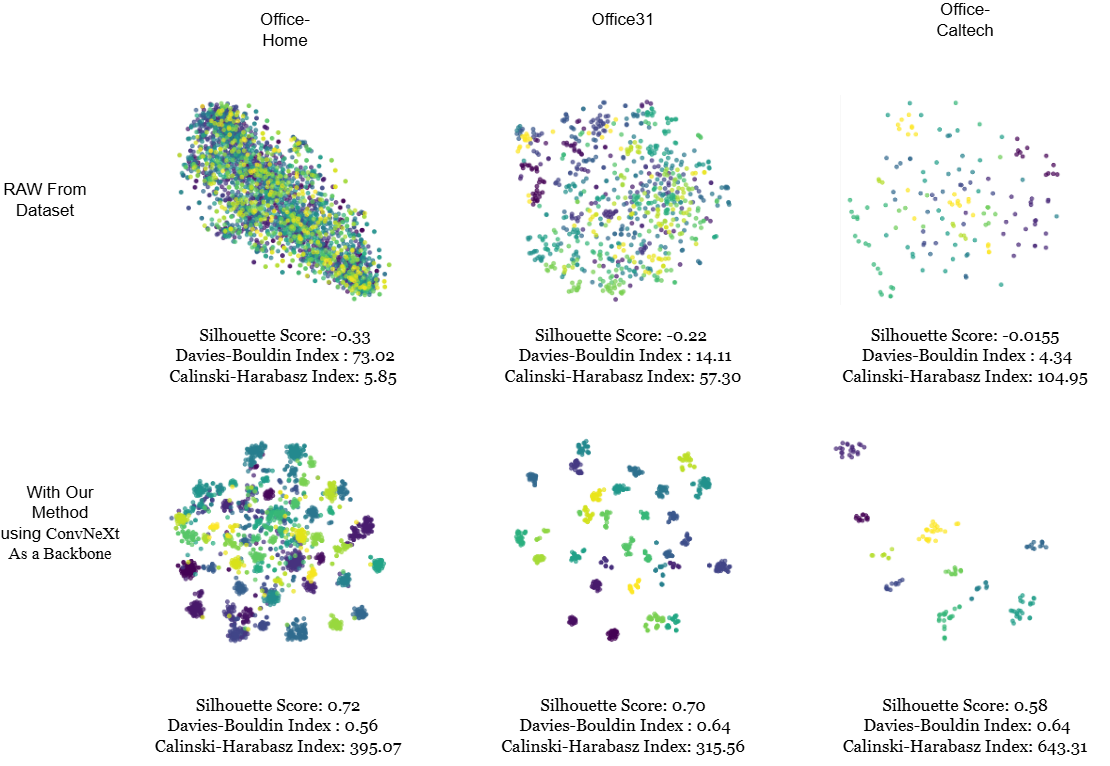} 
\caption{t-SNE map showing clustering results with Silhouette scores, Davies-Bouldin Index (DBI), and Calinski-Harabasz Index (CHI) values for different datasets. The comparison between the raw dataset (top row) and the method using ConvNeXt as a backbone (bottom row) highlights improvements in clustering quality as indicated by higher Silhouette scores and lower DBI and CHI values.}
\label{fig:fig_3}
\end{figure*}
\subsection{t-SNE Visualization}
t-SNE (t-Distributed Stochastic Neighbor Embedding) is a non-linear dimensionality reduction method that is frequently used to visualize high-dimensional data in lower-dimensional spaces like 2D or 3D. The key idea behind t-SNE is to convert pairwise similarities of data points in the high-dimensional space into probabilities and then map these probabilities to a lower-dimensional space while preserving their relationships \cite{sun2023laplacian}. The objective of t-SNE is to minimize the mismatch between the distributions \(P_{ij}\) and \(Q_{ij}\) using the Kullback-Leibler (KL) divergence. The equation is represented by Equation \eqref{eq:eq_7}:
\begin{equation}
    \mathcal{L} = \sum_{i}^{}\mspace{2mu}\sum_{j}^{}\mspace{2mu} P_{ij}\log\frac{P_{ij}}{Q_{ij}}\
    \label{eq:eq_7}
\end{equation}
This loss function ensures that the points in the low-dimensional space retain the neighborhood relationships from the high-dimensional space.

Fig. \ref{fig:fig_3} compares the latent space representations of samples from the three datasets using t-SNE visualization.

The first row represents the raw features directly extracted from the datasets, and the second row shows the visualization using features learned with the proposed domain adaptation method that incorporates the ConvNeXt backbone. Each column represents a dataset. 

For each case, three types of clustering metrics were used: The Silhouette Score, Davies-Bouldin Index (DBI), and Calinski-Harabasz Index (CHI). The Silhouette Score quantifies how well-separated clusters are, with values ranging from -1 to 1, where higher values indicate better-defined and more cohesive clusters. DBI value measures the average similarity between each cluster and the one to which it is most similar, with lower values reflecting compact and well-separated clusters. Lastly, the CHI value evaluates the ratio of between-cluster dispersion, with higher values suggesting that clusters are well-separated and distinct from each other. In the upper row, the raw feature representations show scattered and overlapping clusters, indicating poor class separability and lack of alignment between domains. This highlights the difficulties of domain adaptation. For each dataset, a negative silhouette score suggests that many samples are closer to other clusters than their own; a high DBI value points to weak cluster compactness and separation; and a low CHI value confirms poorly defined clustering structures. The bottom row reveals significant improvements in the latent-space representations achieved by the proposed method. The clusters are well-separated and distinct for each class, demonstrating the model's ability to learn domain-invariant and class-discriminative representations. This improvement is proven by clustering metrics. For each dataset, a positive Silhouette Score indicates stronger intra-cluster cohesion and inter-cluster separation, a lower DBI reflects tighter, more compact clusters, and a high CHI value highlights robust clustering with well-defined boundaries.

Overall, this comparison demonstrates how raw features are inadequate for managing domain transitions and how the suggested approach is better at handling these issues. The technique successfully extracts structured, aligned, and class-discriminative features by using ConvNeXt as a backbone, allowing robust performance across several domains. 

\section{Discussion}
\label{discussion}
Our method introduces a novel approach to source-free domain adaptation (SFDA) by utilizing latent space consistency from multiview augmentation. Traditional domain adaptation methods depend on access to source data or employ adversarial training and pseudo-labeling. However, our method focuses on aligning the latent space representations of augmented target samples by minimizing the Mean Squared Error (MSE) between latent embeddings of different augmentations. The model learns domain-invariant features directly from the target domain. This method provides a robust and effective domain adaptation solution, particularly when source data is unavailable.

The main strength of this method is its source-free setup, which avoids reliance on source data for adaptation. The method bypasses costly pseudolabeling processes and avoids the instability associated with adversarial training. It emphasizes learning stable and domain-agnostic representations using multiview latent consistency, which aligns multiple augmented views of the same target sample in the latent space. The proposed method consistently achieves competitive results on benchmark datasets such as Office-31, Office-Home, and Office-Caltech. The model achieves accuracies of 90. 72\% in Office-31, 84\% in Office-Home, and 97. 12\% in Office-Caltech. For example, the method shows strong generalization and adaptability in challenging transfers such as DSLR to Amazon or Art to Clipart. This performance underscores its ability to extract meaningful features from the target domain without relying on source data or external supervision.
\section{Conclusion}
\label{conc}
In conclusion, the proposed method represents a substantial advance in the field of source-free domain adaptation utilizing latent space consistency. This work represents a paradigm shift in source-free domain adaptation. We introduce a framework focused on maintaining latent space consistency by effectively addressing the fundamental challenges posed by domain shifts. Our method avoids the computational burdens and instabilities associated with traditional techniques. It is the first to integrate multiview augmentations, source-free operation, and computational simplicity into a cohesive framework. It enables strong domain-invariant feature learning without needing source data, pseudo-labeling, or adversarial training, providing a practical, efficient, and highly effective solution to real-world domain adaptation challenges.

However, there are some challenges and limitations to our method. This approach is heavily based on augmentation pipelines. Improperly designed augmentations could reduce their effectiveness. The design of adaptive augmentation strategies relevant to the target domain could further enhance the robustness of the method. Additionally, the method lacks mechanisms for directly fine-tuning target labels, which may restrict its usefulness in situations where partial labeling in the target domain is feasible. Lastly, using multiple augmented views is computationally efficient compared to adversarial methods, but it may face scalability challenges when dealing with very large datasets or high-dimensional feature spaces. Using multiple augmented views is computationally efficient because it focuses on generating diverse representations of the data rather than relying on complex adversarial training. Adversarial methods typically involve training additional networks (e.g., a generator and a discriminator) and iteratively optimizing between them, which is computationally expensive and requires substantial resources. However, the scalability challenge arises in this method because, as the dataset size or feature dimensionality grows, managing and processing a large number of augmented views becomes resource-intensive.

Future work could address these limitations by exploring adaptive augmentation pipelines that dynamically adjust to the characteristics of the target domain\cite{zhang2019bridging}. For example, using generative models to create augmentations that are relevant to domain-specific transformations could enhance performance and results \cite{shivashankar2023semantic}. However, few-shot learning methods could enable it to enhance small amounts of labeled target data, providing a more comprehensive solution that could address unsupervised and semi-supervised paradigms. Introducing mechanisms to manage noisy or outlier samples in the target domain could enhance robustness, especially in scenarios where the distributions of datasets are imbalanced.

\section*{Acknowledgments}
The authors declare that they have no financial conflicts of interest that could have influenced this work.

\medskip


\end{document}